\documentclass[11pt]{article}

\usepackage[preprint]{acl}

\usepackage{amsfonts}
\usepackage{times}
\usepackage{latexsym}
\usepackage{graphicx}
\usepackage{subcaption}
\usepackage{float}
\usepackage{amsmath}
\usepackage{algorithm}
\usepackage{algorithmic}
\usepackage{bm}

\usepackage{multirow}
\usepackage[T1]{fontenc}

\usepackage[utf8]{inputenc}

\usepackage{microtype}

\usepackage{inconsolata}

\usepackage{graphicx}

\title{TapOut: A Bandit-Based Approach to Dynamic Speculative Decoding}

\author{
 \textbf{Aditya Sridhar\textsuperscript{1,2}\thanks{Work done while on an internship at Cerebras Systems.}},
 \textbf{Nish Sinnadurai\textsuperscript{1}},
 \textbf{Sean Lie\textsuperscript{1}},
 \textbf{Vithursan Thangarasa\textsuperscript{1}},
\\
\\
 \textsuperscript{1}Cerebras Systems,
 \textsuperscript{2}University of Waterloo
\\
 \small{
   \textbf{Correspondence:} \href{ma27sridh@uwaterloo.ca}{a27sridh@uwaterloo.ca},
   \href{mailto:vithu@cerebras.net}{vithu@cerebras.net}
 }
}

\begin{document}
\maketitle
\begin{abstract}
Speculative decoding accelerates LLMs by using a lightweight draft model to generate tokens autoregressively before verifying them in parallel with a larger target model. However, determining the optimal number of tokens to draft remains a key challenge limiting the approach's effectiveness. Dynamic speculative decoding aims to intelligently decide~\textit{how many} tokens to draft to achieve maximum speedups. Existing methods often rely on hand-tuned, sensitive thresholds (e.g., token entropy), which are costly to set and generalize poorly across models and domains. We propose TapOut, an~\textit{online, training-free, plug-and-play} algorithm for dynamic speculation policy selection using multi-armed bandits. Our approach employs a meta-algorithm that selects among multiple parameter-free dynamic speculation strategies based on past reward and exploration. We conduct extensive experiments across diverse model pairs and datasets, showing that TapOut achieves competitive or superior speedups compared to well-established dynamic speculation baselines~\textit{without any hyperparameter tuning}. 
\end{abstract}


\section{Introduction}

Large Language Models generate text in an autoregressive manner, where the generation of each token depends on the entire preceding context. Speculative decoding~\cite{10.5555/3618408.3619203, chen2023accelerating} accelerates decoder-only models by using a cheap draft model to generate tokens autoregressively while a larger target model performs verification of these tokens in parallel. This approach leads to massive gains in throughput as target model forward passes are minimized. 

\begin{figure}[t]
    \centering
    \includegraphics[width=.9\linewidth]{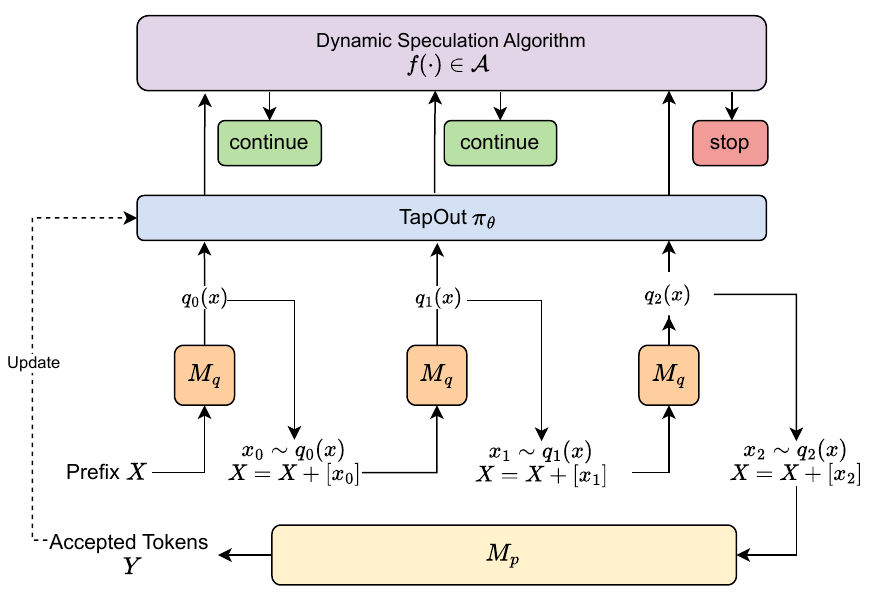}
    \caption{Dynamic speculative decoding with TapOut. After the draft model \(M_q\) generates predictions for the next token, TapOut \(\pi_\theta\) uses a bandit algorithm to select a dynamic speculation algorithm \(f(\cdot)\) which returns a decision to stop or continue drafting. If \(f(\cdot)\) chooses to stop drafting, the target model \(M_p\) performs verification of all generated tokens and selects a subset that match its predictions. The accepted tokens are used to update the bandit policy.}
    \label{fig:diagram}
\end{figure}

Ideally, drafting should continue until the next token is not accepted by the target model for maximum speedups. However, most previous works~\cite{li2024eagle, cai2024medusa, he2024restretrievalbasedspeculativedecoding, fu2024breaksequentialdependencyllm} rely on a static draft length. Since oracle draft length exhibits high variance~\cite{agrawal2024adaedl, mamou2024accelerating}, a static draft length is inherently suboptimal and results in missed speedup opportunities.


Previous dynamic speculation techniques have sought to address this challenge. \textit{Training-free} methods propose entropy-based stopping criteria~\cite{zhang2024draft, agrawal2024adaedl}, while \textit{training-based} methods train a classifier model to make probabilistic stopping decisions~\cite{mamou2024accelerating, huang2024specdec++}. However, these techniques introduce hyperparameters, such as thresholds, which are costly to tune. 
Furthermore, none of them act in an~\textit{online} manner and thus cannot adapt to shifts in the prompt distribution.

Recently, BanditSpec~\cite{hou2025banditspec}, reframed speculative decoding hyperparameter selection as a multi-armed bandit (MAB) problem. However, our work is distinct: BanditSpec aims to perform~\textbf{static speculative decoding configuration selection} (e.g., fixed draft length, specific draft model) for the full generation process. In contrast, our method addresses \textbf{the selection of the dynamic stopping policy~} and uses a bandit to make token-wise stop/continue decisions~\textit{within a single draft}.
In this work, we design TapOut, an online dynamic speculation algorithm based on MABs. Our method selects amongst~\textit{parameter- and training-free dynamic speculation techniques} and is thus extremely cheap to deploy. In addition, the application of MABs provides strong interpretability by tracking online parameters, such as arm values. Our contributions can be summarized as follows.
\begin{itemize}
    \item We cast dynamic stopping in speculative decoding as an online decision problem over training-free policies, using a multi-armed bandit (MAB) to make token/sequence-local stop/continue decisions within each draft.
    
    \item We instantiate Upper Confidence Bound (UCB) and Thompson Sampling (TS) bandits over a pool of parameter- and training-free dynamic speculation rules and define a blended reward that aligns with speedup by balancing acceptance length and rate; the bandit’s arm values provide an interpretable, online readout of which heuristic is best \textit{now}.
    
    \item Across diverse model pairs and datasets, our sequence-level UCB1 achieves competitive or superior speedups to tuning-required baselines—without any hyperparameter search; learned arm values track baseline ordering across prompts, evidencing faithful, context-aware adaptation.
    
\end{itemize}

\section{Related Work}
Speculative decoding algorithms~\cite{10.5555/3618408.3619203, chen2023accelerating} have been well-studied and several different approaches have been proposed for drafting and verification. EAGLE~\cite{li2024eagle} and related work~\citep{cai2024medusa} eliminate the use of a separate draft model and modify the target model itself to support cheap approximations of the full forward pass. Other approaches, such as REST ~\citep{he2024restretrievalbasedspeculativedecoding}, use simple data structures, like Tries, instead of neural networks as candidate sequence generators. We refer the interested reader to~\citep{xia2024unlockingefficiencylargelanguage} for a comprehensive survey and comparison of popular techniques. These techniques largely cover \textit{how} to draft and verify tokens.

Dynamic speculative decoding is a much more recent branch of literature concerning \textit{how long} to draft tokens before performing verification. Early works can largely be categorized as \textit{training-based} and \textit{training-free}. Training-based works~\citep{huang2024specdec++, mamou2024accelerating} train a classifier to decide whether to keep drafting tokens. Training-free works either use token-level signals, such as entropy~\citep{zhang2024draft, agrawal2024adaedl}, to make stopping decisions or completely modify the speculative decoding paradigm to run draft and target models in parallel~\citep{liu2024parallel, shen2025speculative}. In this work, we focus on training-free dynamic speculation strategies, which use token-level signals as they are simple to implement in existing speculative decoding pipelines, but suffer from sensitive thresholds which limit transferability across model families and datasets.

\section{Methodology}
This section formalizes dynamic stopping in speculative decoding as an online decision problem. We introduce TapOut, a training-free bandit controller that selects among uncertainty-based policies to maximize performance.

\subsection{Problem Setup}
A Large Language Model (LLM) defines a probability distribution over sequences of tokens. Given a context sequence \(x_{<t} = (x_1, x_2, \dots, x_{t-1})\), the model autoregressively estimates the probability of the next token \(x_t\) via \(P(x_t \mid x_{<t}; \theta)\), where \(\theta\) denotes the model parameters.
In speculative decoding, a lightweight draft model first proposes a sequence of tokens \(
X = (x_1, x_2, \dots, x_\gamma)\),
which are then verified in parallel by a more accurate target model. The target model accepts a prefix of \(X\), denoted \(
Y = (y_1, y_2, \dots, y_m), \text{where } m \leq \gamma \text{ and } Y \subseteq X.
\)


A multi-armed bandit problem is defined over a 2-tuple (\(\mathcal{A}, \mathcal{R}\)). Here, \(\mathcal{A}\) is the set of actions that the agent with policy \(\pi\) can take at each timestep, and \(\mathcal{R}^a\) is the stationary reward distribution for action \(a \in \mathcal{A}\), with expected reward \(\mathbb{E}[R^a] = \mathbb{P}(r \mid a)\). For a finite horizon of time steps \(T\), the agent aims to maximize its cumulative reward, \(\sum_{t=1}^{T} r_t\), where \(r_t\sim R^{a_t}\) is the reward received at time step \(t\) based on the action taken, \(a_t\).
In our case, each \(a \in \mathcal{A}\) is a dynamic speculation technique, represented as a function \(f(\cdot)\) that returns a decision \(d \in \{\texttt{stop, continue}\}\). 

We consider two setups with different action granularity: 1)~\textit{sequence-level} bandits where actions are chosen at the beginning of each drafting session and all generation steps use the same dynamic speculation technique, and 2)~\textit{token-level} bandits where each position in the drafting sequence is itself an MAB. 

\subsection{Reward Formulation}
\label{sec:reward}

The agent must manage a core trade-off between speculation aggressiveness and efficiency. Maximizing the acceptance length requires drafting longer, more speculative sequences, while maximizing the acceptance rate requires drafting shorter, higher-confidence sequences to avoid rejection.

We test two reward formulations for the sequence-level bandit, \(r^{simple}\) and \(r^{blend}\), where \(r^{simple} = \frac{|Y|}{\gamma}\) and \(\quad
r^{blend} = \alpha \cdot \frac{|Y|}{\gamma} + (1-\alpha) \cdot \frac{|Y|}{|X|}\).
Note that \(r^{simple}\) is the normalized acceptance length and \(r^{blend}\) blends both normalized acceptance length and acceptance rate. Speedup is a direct outcome of maximizing acceptance length and rate, and thus \(r^{blend}\) serves as an effective proxy reward. In practice, we fix \(\alpha =\) 0.5 for equal weighting on the two objectives. For the token-level bandit, the reward is simply  \(r = \) 1 if the token is accepted and \(r =\)  0 otherwise.

\subsection{TapOut}

Figure~\ref{fig:diagram} demonstrates the overall flow while abstracting the specific bandit algorithm \(\pi_{\theta}\) used. After the draft model generates logits for the next token, TapOut is queried and selects a dynamic speculation algorithm which returns a stopping decision. If the algorithm chooses to stop, the target model performs verification of the candidate tokens and the bandit's parameters are updated. Algorithmic pseudocode is provided in Algorithm~\ref{alg:tapout}.

\label{sec:pseudocode}
\begin{algorithm}[t]
\caption{TapOut}
\label{alg:tapout}
\begin{algorithmic}[1]
\REQUIRE Draft model \(M_q\), Target model \(M_p\), Bandit \(\pi_\theta\), Prefix \texttt{prefix}, Max draft length \(\gamma\)

\STATE Initialize draft sequence \(X \leftarrow []\)

\FOR{\(i = 1\) to \(\gamma\)}
    \STATE Sample token \(x_i \sim M_q(\texttt{prefix} + X)\)
    \STATE Append \(x_i\) to \(X\)
    
    \IF{\(\pi_\theta\) signals stop at step \(i\)}
        \STATE \textbf{break}
    \ENDIF
\ENDFOR

\STATE Verify \(X\) with \(M_p\), accept maximal matching prefix \(Y\)

\STATE Sample final token \(x_t \sim M_p(\texttt{prefix} + Y)\)

\STATE Update bandit: \texttt{update}(\(\pi_\theta, |Y|, |X|, \gamma\))

\RETURN \(\texttt{prefix} + Y + [x_t]\)

\end{algorithmic}
\end{algorithm}



We implement Upper Confidence Bound (UCB) and Thompson Sampling algorithms. 
For UCB, we compare UCB1~\cite{auer2002finite} and UCB-Tuned algorithms. At time step \(t\), UCB1 selects the arm \(a\) maximizing the sum of the empirical mean reward and an exploration bonus:
\begin{align*}
    a_t = \arg\max_{a \in \mathcal{A}} \left( \hat{\mu}_a(t) + \sqrt{\frac{2 \log t}{N_a(t)}} \right),
\end{align*} where \(
\hat{\mu}_a(t) = \frac{1}{N_a(t)} \sum_{s=1}^{t-1} r_s \mathbf{1}_{\{a_s = a\}} \) is the empirical mean reward of arm \(a\), and \(N_a(t)\) is the number of times arm \(a\) has been played before time \(t\).

UCB-Tuned refines the exploration process by incorporating variance estimates for each arm's rewards to incentivize early exploitation for low-variance arms while retaining a larger exploration bonus for high variance arms:
\begin{equation*}
\begin{aligned}
a_t &= \arg\max_{a \in \mathcal{A}} \Bigg(
  \hat{\mu}_a(t) \\
&\quad + \sqrt{\frac{\log t}{N_a(t)} \min\!\left(\tfrac{1}{4}, V_a(t)\right)}
\Bigg),
\end{aligned}
\end{equation*}
where the variance estimate is:
\[
V_a(t) = \hat{\sigma}^2_a(t) + \sqrt{\frac{2 \log t}{N_a(t)}},
\]
and \(\hat{\sigma}^2_a(t)\) is the empirical variance of the rewards observed for arm \(a\).

Thompson Sampling (TS) maintains a posterior distribution over the expected reward \(\theta_a\) for \(a \in \mathcal{A}\). At each timestep \(t\), it samples a reward estimate from the posterior, \(\tilde{\theta}_a(t) \sim p(\theta_a \mid \mathcal{D}_t)\), where \(\mathcal{D}_t\) is the history of observed actions and rewards up to \(t\). The action selected is \(a_t = \arg\max_{a \in \mathcal{A}} \tilde{\theta}_a(t)\). After playing arm \(a_t\) and observing reward \(r_t\), the posterior is updated via Bayes' rule, \(
p(\theta_{a_t} \mid \mathcal{D}_{t+1}) \propto p(r_t \mid \theta_{a_t}) \, p(\theta_{a_t} \mid \mathcal{D}_t)\).
The token-level problem provides token-wise binary rewards for each bandit, and thus we can use the standard Beta-Bernoulli Thompson Sampling configuration.
In contrast, the sequence-level problem uses a continuous reward \(r_t \in [0, 1]\)  representing the quality of an entire generated sequence, so here we use a Gaussian prior with a known noise variance. 





\section{Experiments}
We evaluate acceptance length \(m\), acceptance rate \(\%\), and speedup \(s\) over vanilla speculative decoding with a fixed draft length of \(\gamma=6\). For dynamic speculation strategies, the max draft length is set to \(128\) tokens as a proxy for unbounded draft lengths.
Table~\ref{tab:arm_algorithms} describes the bandit's arm algorithms with set thresholds. Note that these are fixed and not tuned on any dataset. SVIPDifference and LogitMargin are new techniques which we describe in Appendix~\ref{sec:arms}.

\begin{table}[t]
  \centering
  \caption{TapOut arm algorithms with stopping condition and, if applicable, threshold. \(\hat{x}_1\) and \(\hat{x}_2\) represent the top-1 and top-2 logit indices respectively.}
  \resizebox{\linewidth}{!}{%
    \begin{tabular}{lcc}
      \hline
      \textbf{Algorithm} & \textbf{Stopping Condition} & \textbf{Threshold \(h\)} \\
      \hline
      Max-Confidence & \( p(x_t = \hat{x}_1 \mid x_{<t}) < h \) & 0.8  \\
      SVIP~\citep{zhang2024draft}  & \( \sqrt{\mathcal{H}(p(x_t \mid x_{<t}))} > h \) & 0.6 \\
      AdaEDL~\citep{agrawal2024adaedl}        & \( 1 - \sqrt{\gamma \cdot \mathcal{H}(p(x_t \mid x_{<t}))} < \lambda_t \) & - \\
      SVIPDifference         & \( \sqrt{\mathcal{H}(p(x_t \mid x_{<t}))} - \sqrt{\mathcal{H}(p(x_{t-1} \mid x_{<t-1}))} > h \) & 0.2 \\
      LogitMargin         & \( p(\hat{x}_1 \mid x_{<t}) - p(\hat{x}_2 \mid x_{<t}) < h \) & 0.2 \\
      \hline
    \end{tabular}%
  }
  \label{tab:arm_algorithms}
\end{table}

\subsection{Ablations}
\label{sec:ablations}
We run all ablations on the Llama3.2-1B/3.1 8B pair using SpecBench~\citep{xia2024unlockingefficiencylargelanguage} to study three design choices. First, we analyze entropy at accepted token positions and find that a single global threshold is suboptimal, which motivates an online controller. Second, we compare two reward signals for the sequence-level bandit and adopt a blended reward that produces higher acceptance rates and larger speedups. Third, we compare UCB1 with UCB-Tuned and find that UCB1 achieves a higher speedup across categories. We attribute this to the lower variance of the blended reward, which reduces the benefit of variance-aware exploration and favors the simpler strategy of UCB1.


\subsubsection{Entropy Thresholding}
We provide a practical motivation for our method by analyzing draft model entropy at accepted positions for coding and non-coding prompts. The experimental setup consists of Llama-3 1B/8B running on the SpecBench dataset~\cite{xia2024unlockingefficiencylargelanguage}.

\begin{figure}[tp]
    \centering

    \begin{subfigure}{\linewidth}
        \centering
        \includegraphics[width=\linewidth]{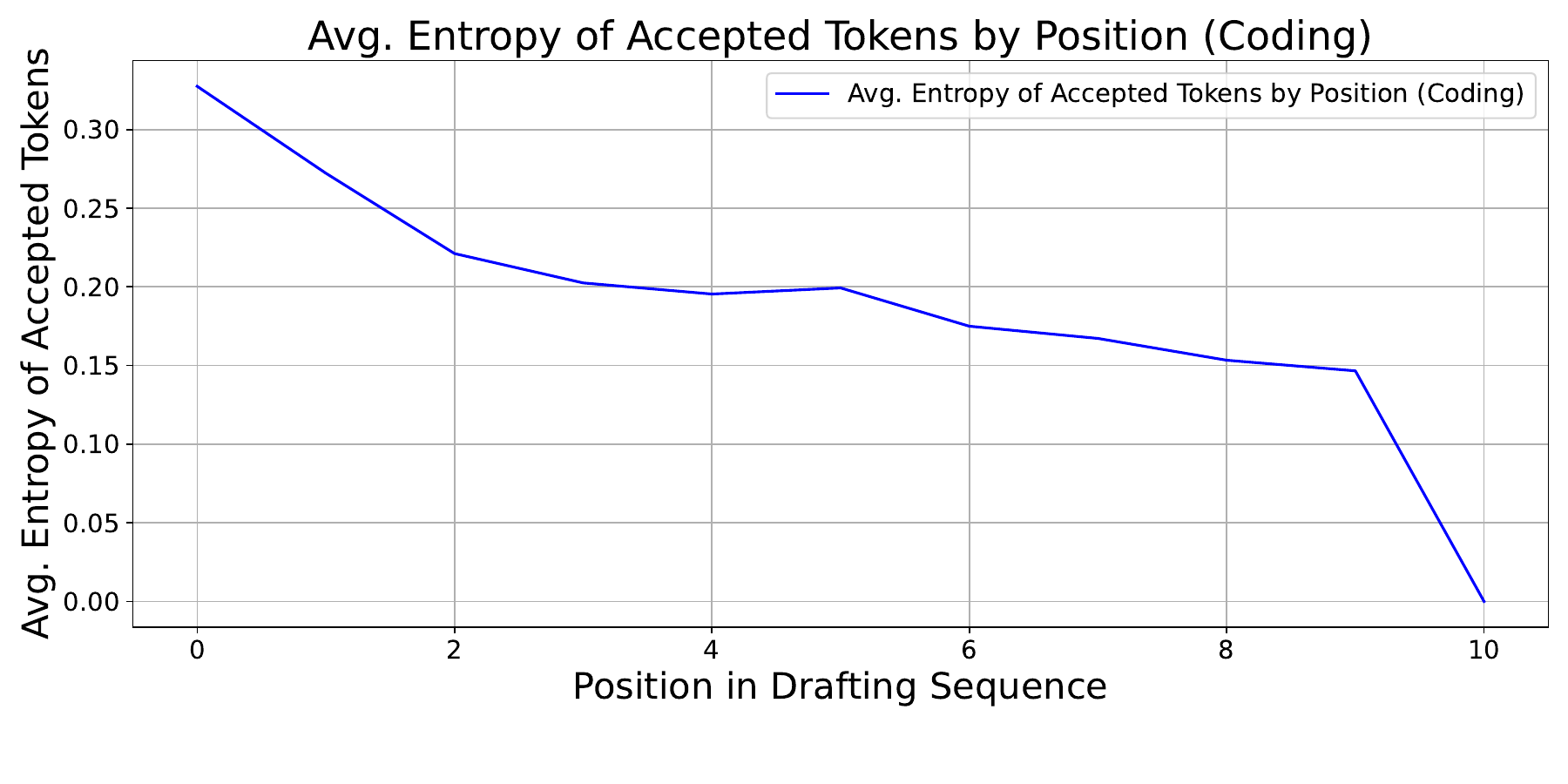}
        \caption{Coding prompts.}
        \label{fig:entropy_tracker_code}
    \end{subfigure}

    \vspace{1em}

    \begin{subfigure}{\linewidth}
        \centering
        \includegraphics[width=\linewidth]{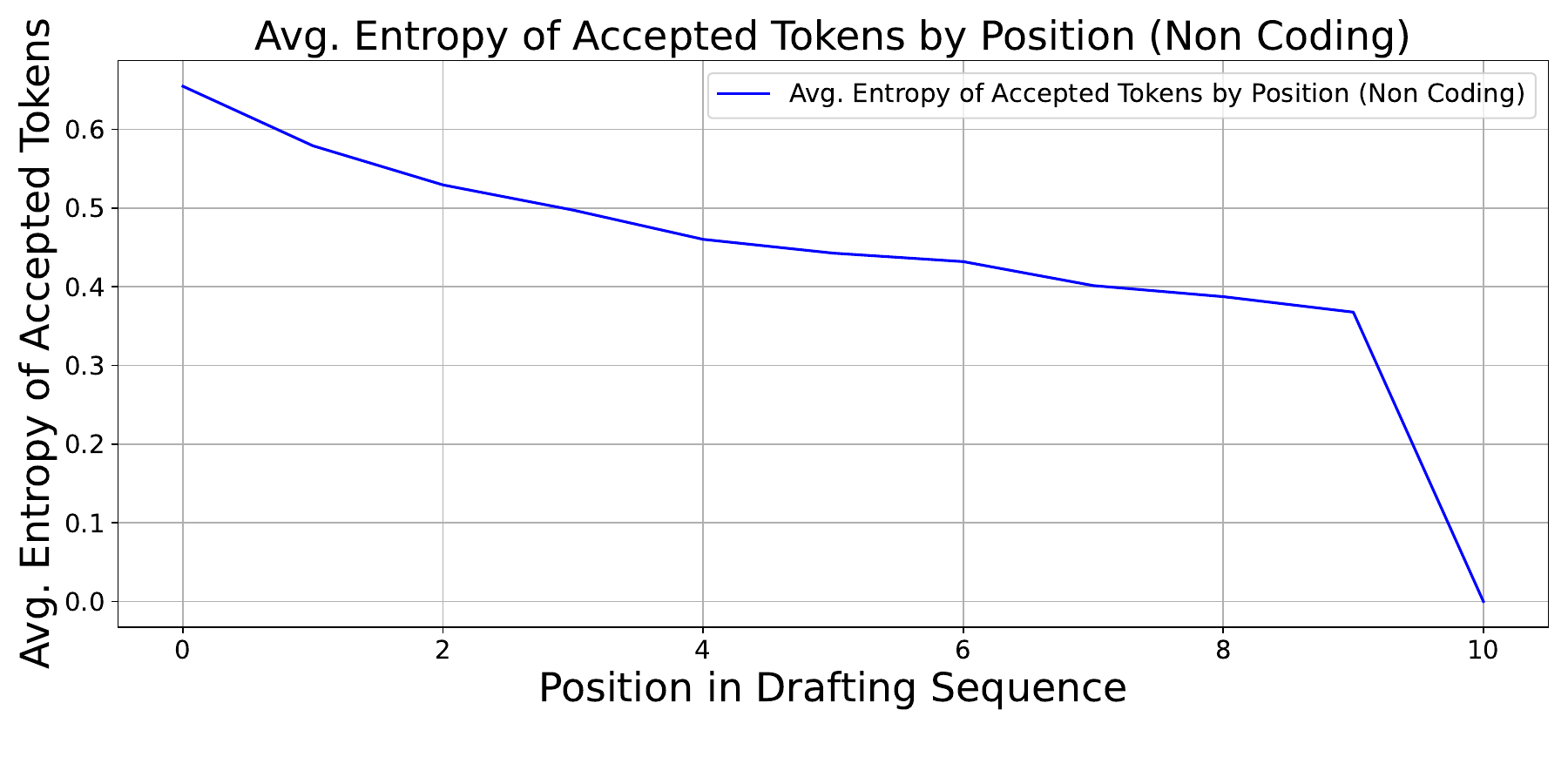}
        \caption{Non-coding prompts.}
        \label{fig:entropy_tracker_non_code}
    \end{subfigure}

    \caption{Draft model \(\sqrt{\mathcal{H}(p(x_t \mid x_{<t}))}\) by position \(t\) for accepted tokens in responses to coding and non-coding prompts.}
    \label{fig:entropy_comparison}
\end{figure}
As shown in Figure \ref{fig:entropy_comparison}, 1) coding prompts cause significantly lower entropy levels than non-coding prompts and 2) entropy decays with generation length as the draft model becomes more confident. Thus, a static threshold across all prompts and positions in the drafting sequence, such as the one used in SVIP~\citep{zhang2024draft}, is suboptimal. An online method is necessary to adapt on-the-fly to changes in the context and possibly favor a different technique.

\subsubsection{Reward Formulation}
\label{sec:reward_ablation}






\begin{figure}[t]
    \centering
    \includegraphics[width=\linewidth]{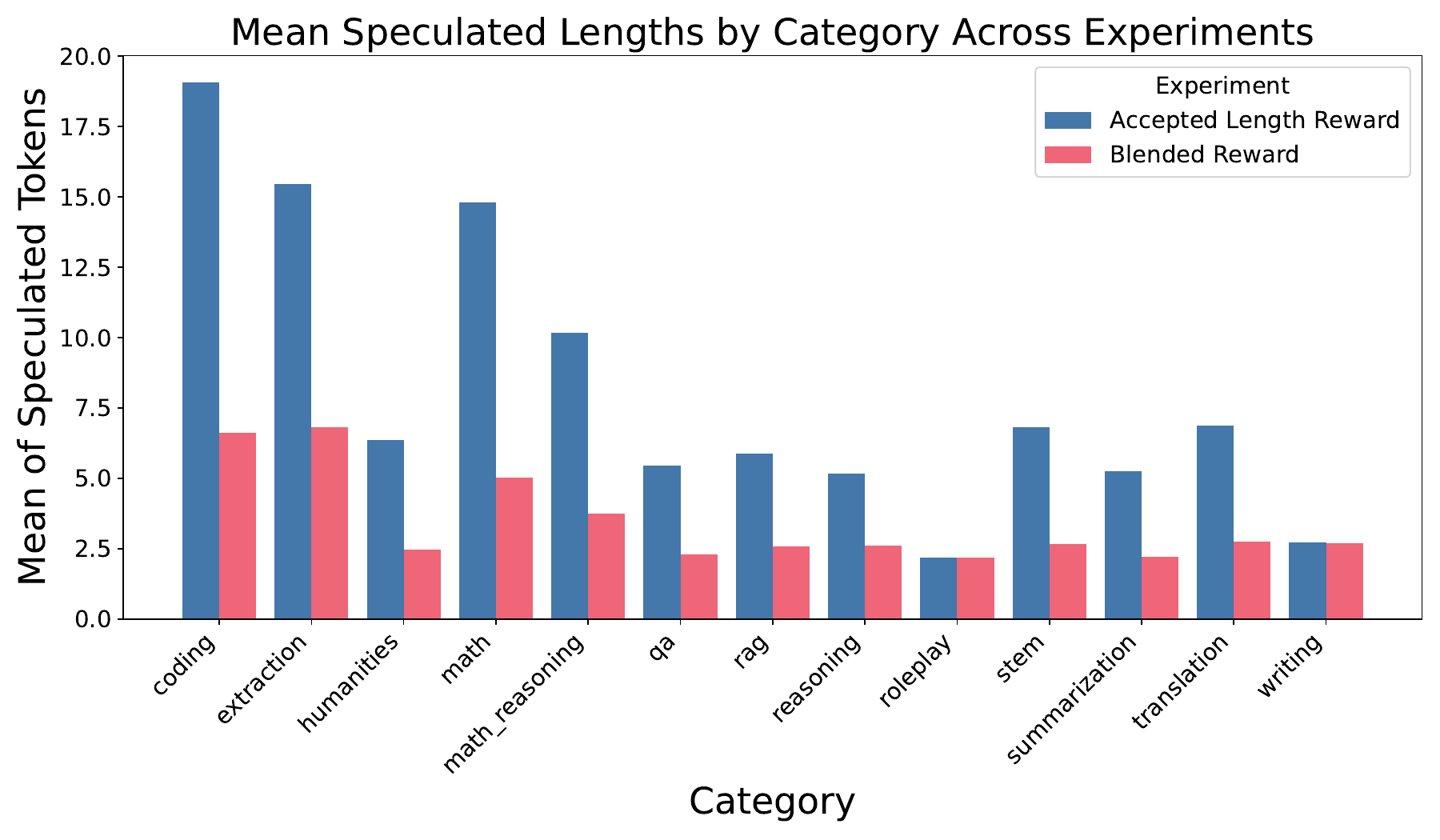}
    \caption{Comparison of speculated length across reward types. The accepted length reward (\(r^{simple}\)) causes the agent to aggressively speculate while the blended reward (\(r^{blend}\)) acts more conservatively. }
    \label{fig:reward_spec_lengths}
\end{figure}


\label{sec:reward_comparison_by_category}
\begin{table}[t]
  \centering
 \caption{Comparison of accepted length and speedup across reward types (\(r^{simple}\) and \(r^{blend}\), see Section~\ref{sec:reward}) for the UCB1 bandit on SpecBench. We find \(r^{blend}\) outperforms \(r^{simple}\) in all categories.}
  \resizebox{0.8\columnwidth}{!}{
    \begin{tabular}{lcccccccc}
      \hline
      \textbf{Category} & \multicolumn{2}{c}{$r^{simple}$} & \multicolumn{2}{c}{$r^{blend}$} \\
      \cline{2-3} \cline{4-5}
                        & $\%$ & $s$ & $\%$  & $s$ \\
      \hline
      coding               & 0.48 & 0.78 & \textbf{0.83} & \textbf{1.18} \\
      extraction           & 0.48 & 0.78 & \textbf{0.79} & \textbf{1.09} \\
      humanities           & 0.51 & 0.99 & \textbf{0.78} & \textbf{1.23} \\
      math                 & 0.53 & 0.82 & \textbf{0.86} & \textbf{1.13} \\
      math reasoning      & 0.57 & 0.89 & \textbf{0.87} & \textbf{1.13} \\
      qa                   & 0.54 & 1.07 & \textbf{0.79} & \textbf{1.27} \\
      rag                  & 0.50 & 1.04 & \textbf{0.79} & \textbf{1.22} \\
      reasoning            & 0.59 & 1.11 & \textbf{0.79} & \textbf{1.26} \\
      roleplay             & \textbf{0.77} & \textbf{1.28} & \textbf{0.77} & \textbf{1.28} \\
      stem                 & 0.52 & 0.95 & \textbf{0.80} & \textbf{1.22} \\
      summarization        & 0.50 & 1.08 & \textbf{0.76} &\textbf{ 1.30} \\
      translation          & 0.42 & 0.99 & \textbf{0.73} & \textbf{1.34} \\
      writing              & 0.79 & 1.25 & \textbf{0.80} & \textbf{1.26} \\
      \hline
    \end{tabular}
  }
  \label{tab:reward_results}
\end{table}

Table~\ref{tab:reward_results} shows the performance of a sequence-level UCB1 bandit for both reward formulations. The blended reward, as described by \(r^{blend}\) in Section \ref{sec:reward}, results in a larger speedup and higher acceptance rates compared to the acceptance length reward, \(r^{simple}\), across all categories. 
This discrepancy arises because the~\(r^{simple}\) reward function is an incomplete proxy for throughput. By rewarding only the total number of accepted tokens while ignoring the acceptance rate, this objective incentivizes the bandit to speculate aggressively, as this strategy maximizes the reward signal even at the cost of many rejected tokens, as shown in Figure~\ref{fig:reward_spec_lengths}.
Thus, we fix the reward to \(r^{blend}\) for all future experiments. 

\subsubsection{UCB1 vs UCB-Tuned}
Next, we ablate the UCB algorithm. Although the oracle draft length exhibits a high variance~\cite{agrawal2024adaedl, mamou2024accelerating}, which provides UCB-Tuned an advantage, we find that the standard UCB1 algorithm provides greater speedup across all categories and thus adopt it for future experiments. We attribute this to the significantly lower variance of the~\(r^{blend}\) reward than~\(r^{simple}\). This stability obviates the need for UCB-Tuned's sophisticated variance-aware exploration, favoring the simpler yet more effective strategy of UCB1.

\begin{figure}[t]
    \centering
    \includegraphics[width=\linewidth]{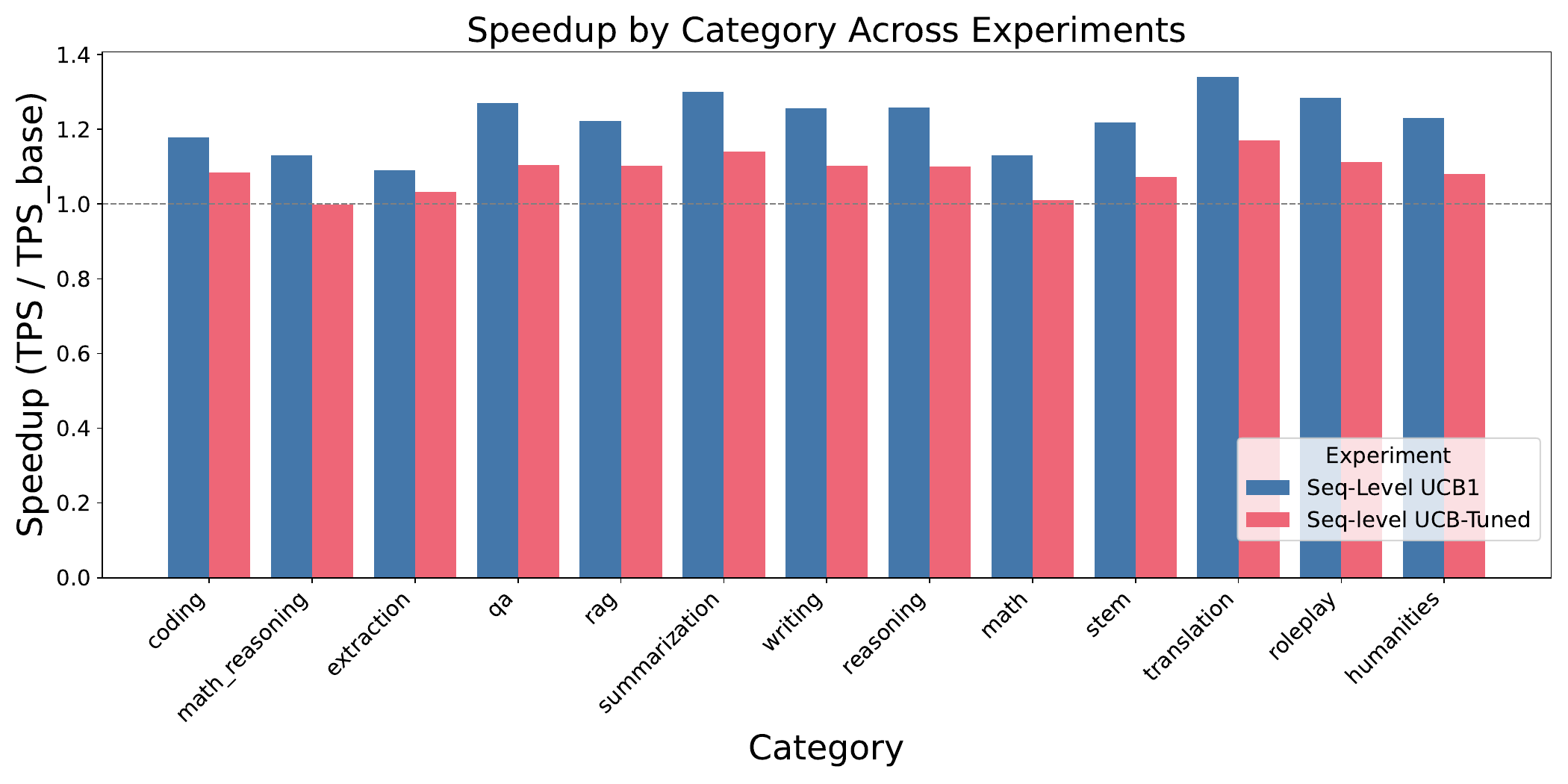}
    \caption{Comparison of speedup between UCB1 and UCB-Tuned. UCB1 provides better performance across all prompt categories.}
    \label{fig:ucb_tuned_speedup}
\end{figure}



\subsection{Results}
\label{sec:results}
We evaluate our method against several existing training-free baselines, including Max-Confidence (MC), SVIP~\cite{zhang2024draft}, and AdaEDL~\cite{agrawal2024adaedl}. We use the Llama~\citep{grattafiori2024Llama3herdmodels} 3.2 1B/3.1 70B, Llama-3.2 1B/3.1 8B, Gemma3~\citep{gemmateam2025gemma3technicalreport} 270M/27B\footnote{Gemma 3 employs sparse attention mechanisms that are not fully optimized in the Transformers library~\cite{wolf-etal-2020-transformers}, which can lead to limited or no speedup in some cases.} and OLMo-2~\cite{olmo20252olmo2furious} 1B/32B model pairs on the MT-Bench~\cite{zheng2023judgingllmasajudgemtbenchchatbot} and HumanEval~\cite{chen2021evaluatinglargelanguagemodels} datasets. Results on SpecBench~\cite{xia2024unlockingefficiencylargelanguage} are also provided in Appendix~\ref{sec:specbench_results}. 

We train a SpecDec++~\cite{huang2024specdec++} classifier on 40,000 samples from the~\texttt{tatsu-lab/alpaca}\footnote{\url{https://huggingface.co/datasets/tatsu-lab/alpaca}} dataset's train split. Also, we use the original SpecDec++ training and inference hyperparameters: BCE rejection weight\,=\ 6, a 4-layer ResNet~\citep{he2015deepresiduallearningimage} with SiLU activation, token-mixing ratio\,=\ 0.15, and stopping threshold\,=\ 0.7. We compare its performance with our bandit methods in Table~\ref{tab:results_specdecpp_comparison}.
For baseline methods with a threshold, we perform a grid search using Llama-3.2 1B/3.1 8B on the SpecBench dataset and fix the optimal values for all other model pairs and datasets.

\begin{table}[t]
  \centering
 \caption{Evaluation of dynamic speculation methods on MT-Bench and HumanEval. \textbf{Bolded} and \textit{italicized} values represent best and second-best results for a given model family and dataset, respectively.}
  \renewcommand{\arraystretch}{1.1}
  \resizebox{\columnwidth}{!}{%
    \begin{tabular}{llc|ccc|ccc}
      \hline
      \textbf{Model} & \textbf{Method} & \textbf{Tuning Required?} &
      \multicolumn{3}{c|}{\textbf{MT-Bench}} &
      \multicolumn{3}{c}{\textbf{HumanEval}} \\
      & & & \textbf{\(m\)} & \textbf{\(\%\)} & \textbf{\(s\)} &
                  \textbf{\(m\)} & \textbf{\(\%\)} & \textbf{\(s\)} \\
      \hline

      \multirow{8}{*}{\shortstack[l]{Llama-3\\1B/70B}}
        & Static-6         & No  & 3.51 & 0.59 & 1.00 & 4.58 & 0.76 & 1.00 \\
        & AdaEDL           & Yes & 2.77 & 0.88 & 0.93 & 4.39 & 0.85 & 1.03 \\
        & SVIP             & Yes & 3.63 & 0.86 & 1.04 & 6.11 & 0.82 & 1.13 \\
        & MC               & Yes & 5.30 & 0.82 & \textbf{1.16} & 8.58 & 0.73 & \textit{1.14} \\
        & TapOut - Seq TS           & No  & 1.04 & 0.78 & 0.76 & 4.12 & 0.84 & 1.04 \\
        & TapOut - Seq UCB1         & No  & 5.29 & 0.82 & \textit{1.15} & 8.48 & 0.73 & \textbf{1.19} \\
        & TapOut - Token TS         & No  & 0.92 & 0.86 & 0.72 & 2.30 & 0.88 & 0.84 \\
        & TapOut - Token UCB1       & No  & 2.65 & 0.82 & 1.03 & 4.38 & 0.84 & 1.06 \\
      \hline

      \multirow{8}{*}{\shortstack[l]{Llama-3\\1B/8B}}
        & Static-6         & No  & 3.47 & 0.58 & 1.00 & 4.35 & 0.72 & 1.00 \\
        & AdaEDL           & Yes & 1.89 & 0.88 & 1.15 & 3.64 & 0.89 & \textit{1.07} \\
        & SVIP             & Yes & 2.44 & 0.87 & \textbf{1.25} & 3.49 & 0.91 & 1.06 \\
        & MC               & Yes  & 3.14 & 0.82 & \textit{1.18} & 4.63 & 0.86 & \textbf{1.09} \\
        & TapOut - Seq TS           & No  & 1.10 & 0.81 & 1.04 & 1.39 & 0.88 & 0.85 \\
        & TapOut - Seq UCB1         & No  & 3.14 & 0.82 & \textbf{1.25} & 4.62 & 0.86 & \textbf{1.09} \\
        & TapOut - Token TS         & No  & 1.89 & 0.80 & 1.15 & 4.63 & 0.86 & \textbf{1.09} \\
        & TapOut - Token UCB1       & No  & 2.24 & 0.83 & 1.11 & 2.96 & 0.88 & 0.97 \\
      \hline

      \multirow{8}{*}{\shortstack[l]{OLMo-2\\1B/32B}}
        & Static-6         & No  & 1.91 & 0.32 & 1.00 & 2.13 & 0.35 & 1.00 \\
        & AdaEDL           & Yes & 0.67 & 0.67 & 1.11 & 0.69 & 0.69 & 1.15 \\
        & SVIP             & Yes & 1.17 & 0.67 & 1.24 & 1.29 & 0.67 & \textit{1.35} \\
        & MC               & Yes & 1.38 & 0.60 & 1.25 & 1.57 & 0.60 & \textbf{1.38} \\
        & TapOut - Seq TS           & No  & 0.80 & 0.62 & 1.12 & 0.76 & 0.67 & 1.16 \\
        & TapOut - Seq UCB1         & No  & 1.37 & 0.59 & \textbf{1.27} & 1.57 & 0.60 & 1.25 \\
        & TapOut - Token TS         & No  & 1.38 & 0.60 & \textit{1.26} & 1.20 & 0.66 & 1.21 \\
        & TapOut - Token UCB1       & No  & 1.28 & 0.64 & 1.25 & 1.53 & 0.62 & 1.33 \\
      \hline

      \multirow{8}{*}{\shortstack[l]{Gemma3\\270M/27B}}
        & Static-6         & No  & 3.81 & 0.63 & \textbf{1.00} & 4.37 & 0.73 & 1.00 \\
        & AdaEDL           & Yes & 0.66 & 0.66 & 0.76 & 4.17 & \textit{0.88} & \textbf{1.52} \\
        & SVIP             & Yes & 1.77 & 0.65 & 0.85 & 4.82 & 0.80 & 1.49 \\
        & MC   & Yes & 2.56 & 0.63 & 0.89 & 7.62 & 0.76 & \textit{1.50} \\
        & TapOut - Seq TS     & No  & 0.73 & 0.63 & 0.77 & 1.25 & 0.91 & 1.05 \\
        & TapOut - Seq UCB1   & No  & 2.62 & 0.63 & \textit{0.90} & 7.85 & 0.77 & \textbf{1.52} \\
        & TapOut - Token TS   & No  & 1.28 & 0.65 & 0.85 & 1.73 & 0.86 & 1.21 \\
        & TapOut - Token UCB1 & No  & 1.62 & 0.56 & 0.81 & 3.42 & 0.81 & 1.32 \\
      \hline
    \end{tabular}
  }
  \label{tab:all_results}
\end{table}

\begin{table}[t]
  \centering
  \caption{Comparison of our training-free bandit techniques to the training-based SpecDec++ using Llama-3 1B/8B on SpecBench.}
  \resizebox{\columnwidth}{!}{%
  \begin{tabular}{llcccc}
    \hline
    \textbf{Model} & \textbf{Method} & \textbf{Training Required?} & \textbf{\(m\)} & \textbf{\(\%\)} & \textbf{\(s\)} \\
    \hline
    \multirow{6}{*}{\shortstack[l]{Llama-3\\1B/8B}} 
        & Static-6        & No & 3.29 & 0.55 & 1.00 \\
        & SpecDec++      & Yes  & 2.73 & 0.65 & 0.99 \\
        & TapOut - Seq TS   & No  & 1.00 & 0.78 & 0.90 \\
        & TapOut - Seq UCB1 & No  & 2.76 & 0.81 & \textbf{1.22} \\
        & TapOut - Token TS  & No  & 2.76 & 0.81 & 1.05 \\
        & TapOut - Token UCB1 & No & 2.23 & 0.82 & \textit{1.13} \\
    \hline
  \end{tabular}
  }
  \label{tab:results_specdecpp_comparison}
\end{table}


\begin{figure}[t]
  \centering

  \begin{subfigure}{\linewidth}
    \centering
    \includegraphics[width=\linewidth]{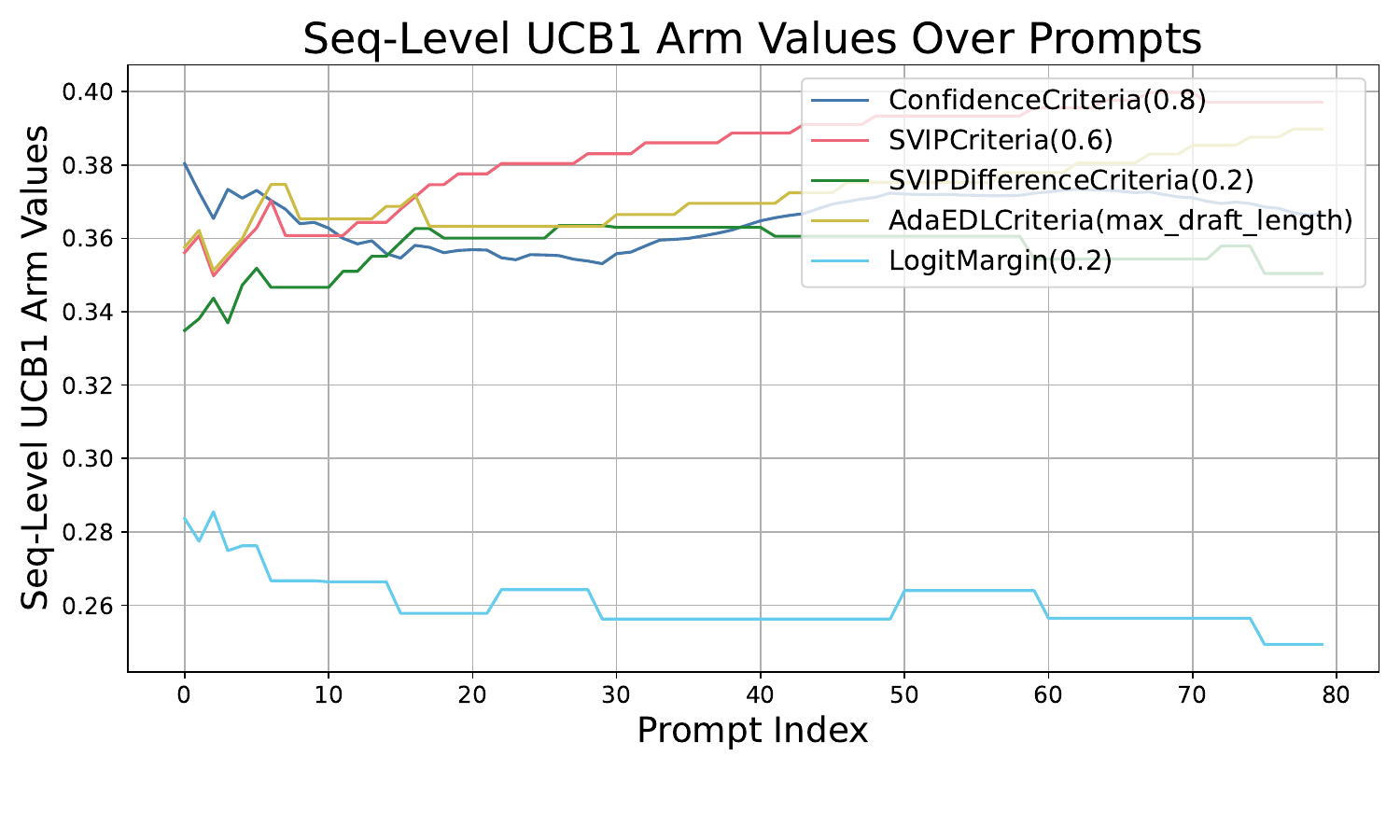}
    \caption{MT-Bench results.}
    \label{fig:acceptance-rate}
  \end{subfigure}

  \vspace{1em} 

  \begin{subfigure}{\linewidth}
    \centering
    \includegraphics[width=\linewidth]{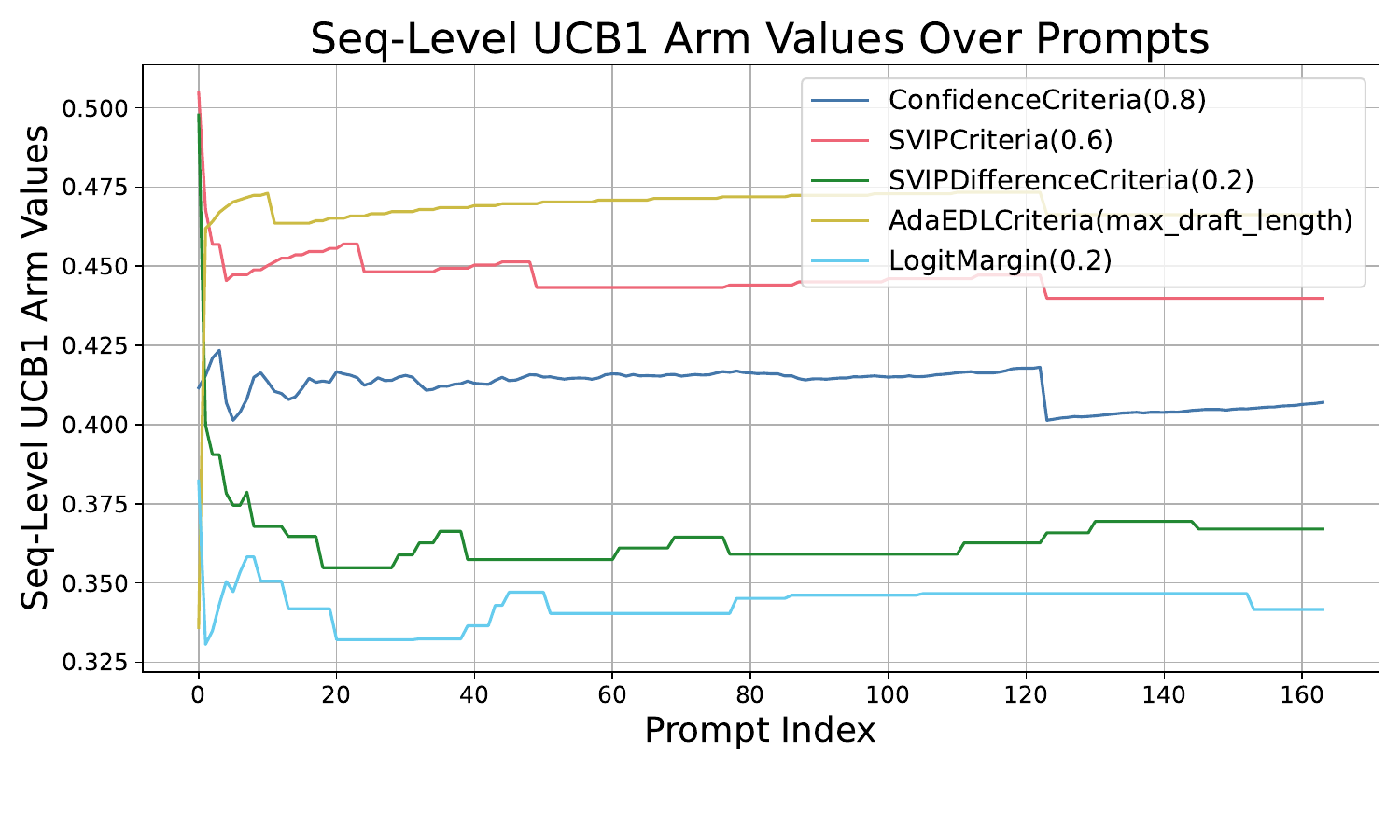}
    \caption{HumanEval results.}
    \label{fig:speedup}
  \end{subfigure}

  \caption{Progression of TapOut Sequence-level UCB1 \(\mu_i\) for Llama-3 1B/8B on a) MT-Bench and b) HumanEval.}
  \label{fig:arm_plots}
\end{figure}

\subsubsection{Analysis}
As shown in Tables~\ref{tab:all_results} and~\ref{tab:results_specdecpp_comparison}, in all except one instance,~\textbf{TapOut sequence-level UCB1 provides top-2 speedup performance while being tuning-free}. This result is particularly noteworthy as the baselines were given a significant advantage: their hyperparameters were tuned directly on the SpecBench benchmark, a process from which TapOut was exempt.
On out-of-distribution tests, such as OLMo-2 on MT-Bench, the adaptive nature of TapOut enables it to outperform these now-suboptimal fixed thresholds, demonstrating the robustness of our approach. Furthermore, when the best baseline varies across datasets (e.g., Llama-3 1B/8B), TapOut matches performance by adaptively choosing stopping policies. 

Token-level bandits likely underperform because each individual bandit receives no feedback on throughput and only binary acceptance/rejection. This can cause the bandit to become over-optimistic, as described in Section~\ref{sec:reward_ablation}. 
\subsection{Interpretability}
\label{sec:arm_plots}

Figure \ref{fig:arm_plots} demonstrates TapOut's interpretability and the effectiveness of the \(r^{blend}\) reward signal. On a) MT-Bench, the bandit quickly identifies SVIP as the dominant strategy and its learned value significantly pulls ahead, with roughly a 0.2 value gap between SVIP and other arms. This mirrors the large performance gap between SVIP and other baseline techniques found in our main results (Table \ref{tab:all_results}). Conversely, on b) HumanEval, where several baselines have close performance, the bandit correctly learns this and the arm values are also tightly clustered, with a 0.025 value gap on average, leading to continued exploration. 

\begin{figure}[tp]
    \centering
    \includegraphics[width=\linewidth]{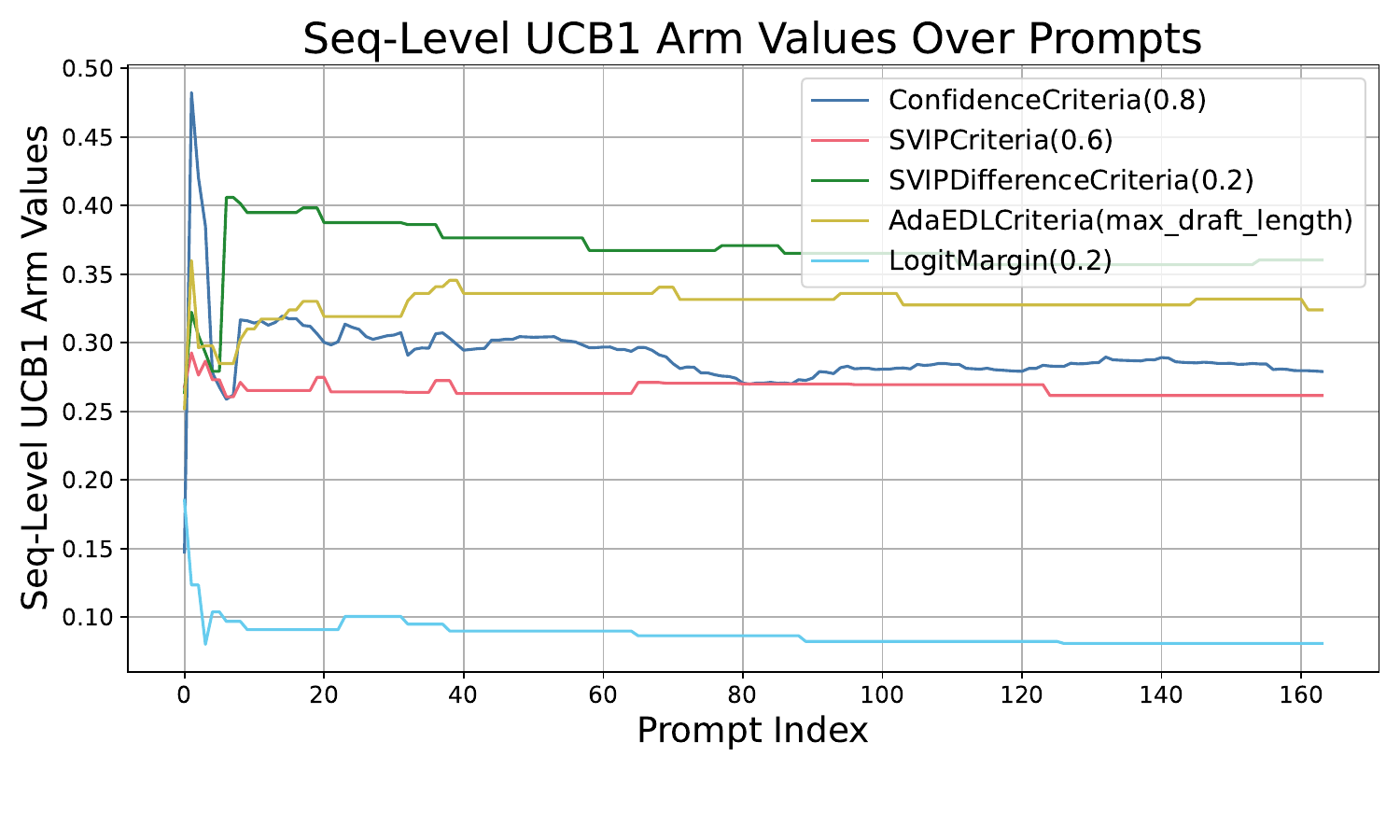}
    \caption{TapOut - Seq UCB1 Arm Value~(\(\mu_i\)) progression for Gemma3 on HumanEval.}
    \label{fig:gemma3_arm_values}
\end{figure}

Together, these results show that the learned arm values accurately reflect the relative performance of the underlying strategies in different contexts. Further, as shown in Figure~\ref{fig:gemma3_arm_values}, the order of arm values exactly matches the order of speedup gains that each baseline obtains in Table~\ref{tab:all_results} for Gemma3 on HumanEval (e.g., AdaEDL > MC > SVIP).

\section{Conclusion}
In this work, we propose TapOut, an online, tuning-free approach to dynamic speculative decoding. Our method uses a MAB to select an algorithm from a pool of training-free dynamic speculation approaches based on past performance, which in turn provides a binary stopping decision. TapOut is simple to implement, highly interpretable, and effectively leverages online feedback to prioritize the best-performing strategies in context. We evaluate our approach using several model families and datasets, demonstrating that the sequence-level UCB1 provides speedups that are competitive or superior to all tuning-required baselines. 

\section{Limitations}
While our work demonstrates the effectiveness and interpretability of bandits for dynamic speculative decoding, there are a few limitations. For instance, the overall performance of bandits does not always exceed the best baseline method, and the datasets we use to evaluate our approach are relatively small. Larger evaluation settings, such as the Alpaca~\citep{alpaca2023} dataset, may provide a better opportunity for online learning. Further, the performance of the bandit is inherently upper-bounded by the individual performances of the dynamic speculation algorithms used as arms. It requires a set of reasonably strong and diverse heuristics in order to make effective online decisions. An interesting follow-up work could investigate other reinforcement learning approaches which leverage context information, such as contextual bandits.

\newpage

\bibliography{custom}

\begin{thebibliography}{23}
\providecommand{\natexlab}[1]{#1}

\bibitem[{Agrawal et~al.(2024)Agrawal, Jeon, and Lee}]{agrawal2024adaedl}
Sudhanshu Agrawal, Wonseok Jeon, and Mingu Lee. 2024.
\newblock \href {https://arxiv.org/abs/2410.18351} {Adaedl: Early draft stopping for speculative decoding of large language models via an entropy-based lower bound on token acceptance probability}.
\newblock \emph{Preprint}, arXiv:2410.18351.

\bibitem[{Auer et~al.(2002)Auer, Cesa-Bianchi, and Fischer}]{auer2002finite}
Peter Auer, Nicolo Cesa-Bianchi, and Paul Fischer. 2002.
\newblock Finite-time analysis of the multiarmed bandit problem.
\newblock \emph{Machine learning}.

\bibitem[{Cai et~al.(2024)Cai, Li, Geng, Peng, Lee, Chen, and Dao}]{cai2024medusa}
Tianle Cai, Yuhong Li, Zhengyang Geng, Hongwu Peng, Jason~D. Lee, Deming Chen, and Tri Dao. 2024.
\newblock \href {https://arxiv.org/abs/2401.10774} {Medusa: Simple llm inference acceleration framework with multiple decoding heads}.
\newblock \emph{Preprint}, arXiv:2401.10774.

\bibitem[{Chen et~al.(2023)Chen, Borgeaud, Irving, Lespiau, Sifre, and Jumper}]{chen2023accelerating}
Charlie Chen, Sebastian Borgeaud, Geoffrey Irving, Jean-Baptiste Lespiau, Laurent Sifre, and John Jumper. 2023.
\newblock \href {https://arxiv.org/abs/2302.01318} {Accelerating large language model decoding with speculative sampling}.
\newblock \emph{Preprint}, arXiv:2302.01318.

\bibitem[{Chen et~al.(2021)Chen, Tworek, Jun, Yuan, de~Oliveira~Pinto, Kaplan, Edwards, Burda, Joseph, Brockman, Ray, Puri, Krueger, Petrov, Khlaaf, Sastry, Mishkin, Chan, Gray, Ryder, Pavlov, Power, Kaiser, Bavarian, Winter, Tillet, Such, Cummings, Plappert, Chantzis, Barnes, Herbert-Voss, Guss, Nichol, Paino, Tezak, Tang, Babuschkin, Balaji, Jain, Saunders, Hesse, Carr, Leike, Achiam, Misra, Morikawa, Radford, Knight, Brundage, Murati, Mayer, Welinder, McGrew, Amodei, McCandlish, Sutskever, and Zaremba}]{chen2021evaluatinglargelanguagemodels}
Mark Chen, Jerry Tworek, Heewoo Jun, Qiming Yuan, Henrique~Ponde de~Oliveira~Pinto, Jared Kaplan, Harri Edwards, Yuri Burda, Nicholas Joseph, Greg Brockman, Alex Ray, Raul Puri, Gretchen Krueger, Michael Petrov, Heidy Khlaaf, Girish Sastry, Pamela Mishkin, Brooke Chan, Scott Gray, and 39 others. 2021.
\newblock \href {https://arxiv.org/abs/2107.03374} {Evaluating large language models trained on code}.
\newblock \emph{Preprint}, arXiv:2107.03374.

\bibitem[{Fu et~al.(2024)Fu, Bailis, Stoica, and Zhang}]{fu2024breaksequentialdependencyllm}
Yichao Fu, Peter Bailis, Ion Stoica, and Hao Zhang. 2024.
\newblock \href {https://arxiv.org/abs/2402.02057} {Break the sequential dependency of llm inference using lookahead decoding}.
\newblock \emph{Preprint}, arXiv:2402.02057.

\bibitem[{Grattafiori et~al.(2024)Grattafiori, Dubey, Jauhri, Pandey, Kadian, Al-Dahle, Letman, Mathur, Schelten, Vaughan, Yang, Fan, Goyal, Hartshorn, Yang, Mitra, Sravankumar, Korenev, Hinsvark, Rao, Zhang, Rodriguez, Gregerson, Spataru, Roziere, Biron, Tang, Chern, Caucheteux, Nayak, Bi, Marra, McConnell, Keller, Touret, Wu, Wong, Ferrer, Nikolaidis, Allonsius, Song, Pintz, Livshits, Wyatt, Esiobu, Choudhary, Mahajan, Garcia-Olano, Perino, Hupkes, Lakomkin, AlBadawy, Lobanova, Dinan, Smith, Radenovic, Guzmán, Zhang, Synnaeve, Lee, Anderson, Thattai, Nail, Mialon, Pang, Cucurell, Nguyen, Korevaar, Xu, Touvron, Zarov, Ibarra, Kloumann, Misra, Evtimov, Zhang, Copet, Lee, Geffert, Vranes, Park, Mahadeokar, Shah, van~der Linde, Billock, Hong, Lee, Fu, Chi, Huang, Liu, Wang, Yu, Bitton, Spisak, Park, Rocca, Johnstun, Saxe, Jia, Alwala, Prasad, Upasani, Plawiak, Li, Heafield, Stone, El-Arini, Iyer, Malik, Chiu, Bhalla, Lakhotia, Rantala-Yeary, van~der Maaten, Chen, Tan, Jenkins, Martin, Madaan, Malo, Blecher,
  Landzaat, de~Oliveira, Muzzi, Pasupuleti, Singh, Paluri, Kardas, Tsimpoukelli, Oldham, Rita, Pavlova, Kambadur, Lewis, Si, Singh, Hassan, Goyal, Torabi, Bashlykov, Bogoychev, Chatterji, Zhang, Duchenne, Çelebi, Alrassy, Zhang, Li, Vasic, Weng, Bhargava, Dubal, Krishnan, Koura, Xu, He, Dong, Srinivasan, Ganapathy, Calderer, Cabral, Stojnic, Raileanu, Maheswari, Girdhar, Patel, Sauvestre, Polidoro, Sumbaly, Taylor, Silva, Hou, Wang, Hosseini, Chennabasappa, Singh, Bell, Kim, Edunov, Nie, Narang, Raparthy, Shen, Wan, Bhosale, Zhang, Vandenhende, Batra, Whitman, Sootla, Collot, Gururangan, Borodinsky, Herman, Fowler, Sheasha, Georgiou, Scialom, Speckbacher, Mihaylov, Xiao, Karn, Goswami, Gupta, Ramanathan, Kerkez, Gonguet, Do, Vogeti, Albiero, Petrovic, Chu, Xiong, Fu, Meers, Martinet, Wang, Wang, Tan, Xia, Xie, Jia, Wang, Goldschlag, Gaur, Babaei, Wen, Song, Zhang, Li, Mao, Coudert, Yan, Chen, Papakipos, Singh, Srivastava, Jain, Kelsey, Shajnfeld, Gangidi, Victoria, Goldstand, Menon, Sharma, Boesenberg,
  Baevski, Feinstein, Kallet, Sangani, Teo, Yunus, Lupu, Alvarado, Caples, Gu, Ho, Poulton, Ryan, Ramchandani, Dong, Franco, Goyal, Saraf, Chowdhury, Gabriel, Bharambe, Eisenman, Yazdan, James, Maurer, Leonhardi, Huang, Loyd, Paola, Paranjape, Liu, Wu, Ni, Hancock, Wasti, Spence, Stojkovic, Gamido, Montalvo, Parker, Burton, Mejia, Liu, Wang, Kim, Zhou, Hu, Chu, Cai, Tindal, Feichtenhofer, Gao, Civin, Beaty, Kreymer, Li, Adkins, Xu, Testuggine, David, Parikh, Liskovich, Foss, Wang, Le, Holland, Dowling, Jamil, Montgomery, Presani, Hahn, Wood, Le, Brinkman, Arcaute, Dunbar, Smothers, Sun, Kreuk, Tian, Kokkinos, Ozgenel, Caggioni, Kanayet, Seide, Florez, Schwarz, Badeer, Swee, Halpern, Herman, Sizov, Guangyi, Zhang, Lakshminarayanan, Inan, Shojanazeri, Zou, Wang, Zha, Habeeb, Rudolph, Suk, Aspegren, Goldman, Zhan, Damlaj, Molybog, Tufanov, Leontiadis, Veliche, Gat, Weissman, Geboski, Kohli, Lam, Asher, Gaya, Marcus, Tang, Chan, Zhen, Reizenstein, Teboul, Zhong, Jin, Yang, Cummings, Carvill, Shepard, McPhie,
  Torres, Ginsburg, Wang, Wu, U, Saxena, Khandelwal, Zand, Matosich, Veeraraghavan, Michelena, Li, Jagadeesh, Huang, Chawla, Huang, Chen, Garg, A, Silva, Bell, Zhang, Guo, Yu, Moshkovich, Wehrstedt, Khabsa, Avalani, Bhatt, Mankus, Hasson, Lennie, Reso, Groshev, Naumov, Lathi, Keneally, Liu, Seltzer, Valko, Restrepo, Patel, Vyatskov, Samvelyan, Clark, Macey, Wang, Hermoso, Metanat, Rastegari, Bansal, Santhanam, Parks, White, Bawa, Singhal, Egebo, Usunier, Mehta, Laptev, Dong, Cheng, Chernoguz, Hart, Salpekar, Kalinli, Kent, Parekh, Saab, Balaji, Rittner, Bontrager, Roux, Dollar, Zvyagina, Ratanchandani, Yuvraj, Liang, Alao, Rodriguez, Ayub, Murthy, Nayani, Mitra, Parthasarathy, Li, Hogan, Battey, Wang, Howes, Rinott, Mehta, Siby, Bondu, Datta, Chugh, Hunt, Dhillon, Sidorov, Pan, Mahajan, Verma, Yamamoto, Ramaswamy, Lindsay, Lindsay, Feng, Lin, Zha, Patil, Shankar, Zhang, Zhang, Wang, Agarwal, Sajuyigbe, Chintala, Max, Chen, Kehoe, Satterfield, Govindaprasad, Gupta, Deng, Cho, Virk, Subramanian, Choudhury,
  Goldman, Remez, Glaser, Best, Koehler, Robinson, Li, Zhang, Matthews, Chou, Shaked, Vontimitta, Ajayi, Montanez, Mohan, Kumar, Mangla, Ionescu, Poenaru, Mihailescu, Ivanov, Li, Wang, Jiang, Bouaziz, Constable, Tang, Wu, Wang, Wu, Gao, Kleinman, Chen, Hu, Jia, Qi, Li, Zhang, Zhang, Adi, Nam, Yu, Wang, Zhao, Hao, Qian, Li, He, Rait, DeVito, Rosnbrick, Wen, Yang, Zhao, and Ma}]{grattafiori2024Llama3herdmodels}
Aaron Grattafiori, Abhimanyu Dubey, Abhinav Jauhri, Abhinav Pandey, Abhishek Kadian, Ahmad Al-Dahle, Aiesha Letman, Akhil Mathur, Alan Schelten, Alex Vaughan, Amy Yang, Angela Fan, Anirudh Goyal, Anthony Hartshorn, Aobo Yang, Archi Mitra, Archie Sravankumar, Artem Korenev, Arthur Hinsvark, and 542 others. 2024.
\newblock \href {https://arxiv.org/abs/2407.21783} {The llama 3 herd of models}.
\newblock \emph{Preprint}, arXiv:2407.21783.

\bibitem[{He et~al.(2015)He, Zhang, Ren, and Sun}]{he2015deepresiduallearningimage}
Kaiming He, Xiangyu Zhang, Shaoqing Ren, and Jian Sun. 2015.
\newblock \href {https://arxiv.org/abs/1512.03385} {Deep residual learning for image recognition}.
\newblock \emph{Preprint}, arXiv:1512.03385.

\bibitem[{He et~al.(2024)He, Zhong, Cai, Lee, and He}]{he2024restretrievalbasedspeculativedecoding}
Zhenyu He, Zexuan Zhong, Tianle Cai, Jason~D. Lee, and Di~He. 2024.
\newblock \href {https://arxiv.org/abs/2311.08252} {Rest: Retrieval-based speculative decoding}.
\newblock \emph{Preprint}, arXiv:2311.08252.

\bibitem[{Hou et~al.(2025)Hou, Zhang, Du, Zhang, Pan, Pang, Du, Tan, and Yang}]{hou2025banditspec}
Yunlong Hou, Fengzhuo Zhang, Cunxiao Du, Xuan Zhang, Jiachun Pan, Tianyu Pang, Chao Du, Vincent Y.~F. Tan, and Zhuoran Yang. 2025.
\newblock \href {https://arxiv.org/abs/2505.15141} {Banditspec: Adaptive speculative decoding via bandit algorithms}.
\newblock \emph{Preprint}, arXiv:2505.15141.

\bibitem[{Huang et~al.(2025)Huang, Guo, and Wang}]{huang2024specdec++}
Kaixuan Huang, Xudong Guo, and Mengdi Wang. 2025.
\newblock \href {https://arxiv.org/abs/2405.19715} {Specdec++: Boosting speculative decoding via adaptive candidate lengths}.
\newblock \emph{Preprint}, arXiv:2405.19715.

\bibitem[{Leviathan et~al.(2023)Leviathan, Kalman, and Matias}]{10.5555/3618408.3619203}
Yaniv Leviathan, Matan Kalman, and Yossi Matias. 2023.
\newblock Fast inference from transformers via speculative decoding.
\newblock In \emph{Proc. of ICML}.

\bibitem[{Li et~al.(2025)Li, Wei, Zhang, and Zhang}]{li2024eagle}
Yuhui Li, Fangyun Wei, Chao Zhang, and Hongyang Zhang. 2025.
\newblock \href {https://arxiv.org/abs/2401.15077} {Eagle: Speculative sampling requires rethinking feature uncertainty}.
\newblock \emph{Preprint}, arXiv:2401.15077.

\bibitem[{Liu et~al.(2025)Liu, Li, Lv, Liu, Zhu, Hu, and Sun}]{liu2024parallel}
Tianyu Liu, Yun Li, Qitan Lv, Kai Liu, Jianchen Zhu, Winston Hu, and Xiao Sun. 2025.
\newblock {PEARL}: Parallel speculative decoding with adaptive draft length.
\newblock In \emph{Proc. of ICLR}.

\bibitem[{Mamou et~al.(2024)Mamou, Pereg, Korat, Berchansky, Timor, Wasserblat, and Schwartz}]{mamou2024accelerating}
Jonathan Mamou, Oren Pereg, Daniel Korat, Moshe Berchansky, Nadav Timor, Moshe Wasserblat, and Roy Schwartz. 2024.
\newblock \href {https://arxiv.org/abs/2405.04304} {Dynamic speculation lookahead accelerates speculative decoding of large language models}.
\newblock \emph{Preprint}, arXiv:2405.04304.

\bibitem[{OLMo et~al.(2025)OLMo, Walsh, Soldaini, Groeneveld, Lo, Arora, Bhagia, Gu, Huang, Jordan, Lambert, Schwenk, Tafjord, Anderson, Atkinson, Brahman, Clark, Dasigi, Dziri, Guerquin, Ivison, Koh, Liu, Malik, Merrill, Miranda, Morrison, Murray, Nam, Pyatkin, Rangapur, Schmitz, Skjonsberg, Wadden, Wilhelm, Wilson, Zettlemoyer, Farhadi, Smith, and Hajishirzi}]{olmo20252olmo2furious}
Team OLMo, Pete Walsh, Luca Soldaini, Dirk Groeneveld, Kyle Lo, Shane Arora, Akshita Bhagia, Yuling Gu, Shengyi Huang, Matt Jordan, Nathan Lambert, Dustin Schwenk, Oyvind Tafjord, Taira Anderson, David Atkinson, Faeze Brahman, Christopher Clark, Pradeep Dasigi, Nouha Dziri, and 21 others. 2025.
\newblock \href {https://arxiv.org/abs/2501.00656} {2 olmo 2 furious}.
\newblock \emph{Preprint}, arXiv:2501.00656.

\bibitem[{Shen et~al.(2025)Shen, Shen, Kong, Liu, Lu, and Wang}]{shen2025speculative}
Yuhao Shen, Junyi Shen, Quan Kong, Tianyu Liu, Yao Lu, and Cong Wang. 2025.
\newblock \href {https://arxiv.org/abs/2506.01979} {Speculative decoding via hybrid drafting and rollback-aware branch parallelism}.
\newblock \emph{Preprint}, arXiv:2506.01979.

\bibitem[{{Stanford Center for Research on Foundation Models}(2023)}]{alpaca2023}
{Stanford Center for Research on Foundation Models}. 2023.
\newblock Alpaca: A strong, replicable instruction-following model.
\newblock \url{https://crfm.stanford.edu/2023/03/13/alpaca.html}.
\newblock Accessed: 2025-10-01.

\bibitem[{Team et~al.(2025)Team, Kamath, Ferret, Pathak, Vieillard, Merhej, Perrin, Matejovicova, Ramé, Rivière, Rouillard, Mesnard, Cideron, bastien Grill, Ramos, Yvinec, Casbon, Pot, Penchev, Liu, Visin, Kenealy, Beyer, Zhai, Tsitsulin, Busa-Fekete, Feng, Sachdeva, Coleman, Gao, Mustafa, Barr, Parisotto, Tian, Eyal, Cherry, Peter, Sinopalnikov, Bhupatiraju, Agarwal, Kazemi, Malkin, Kumar, Vilar, Brusilovsky, Luo, Steiner, Friesen, Sharma, Sharma, Gilady, Goedeckemeyer, Saade, Feng, Kolesnikov, Bendebury, Abdagic, Vadi, György, Pinto, Das, Bapna, Miech, Yang, Paterson, Shenoy, Chakrabarti, Piot, Wu, Shahriari, Petrini, Chen, Lan, Choquette-Choo, Carey, Brick, Deutsch, Eisenbud, Cattle, Cheng, Paparas, Sreepathihalli, Reid, Tran, Zelle, Noland, Huizenga, Kharitonov, Liu, Amirkhanyan, Cameron, Hashemi, Klimczak-Plucińska, Singh, Mehta, Lehri, Hazimeh, Ballantyne, Szpektor, Nardini, Pouget-Abadie, Chan, Stanton, Wieting, Lai, Orbay, Fernandez, Newlan, yeong Ji, Singh, Black, Yu, Hui, Vodrahalli, Greff, Qiu,
  Valentine, Coelho, Ritter, Hoffman, Watson, Chaturvedi, Moynihan, Ma, Babar, Noy, Byrd, Roy, Momchev, Chauhan, Sachdeva, Bunyan, Botarda, Caron, Rubenstein, Culliton, Schmid, Sessa, Xu, Stanczyk, Tafti, Shivanna, Wu, Pan, Rokni, Willoughby, Vallu, Mullins, Jerome, Smoot, Girgin, Iqbal, Reddy, Sheth, Põder, Bhatnagar, Panyam, Eiger, Zhang, Liu, Yacovone, Liechty, Kalra, Evci, Misra, Roseberry, Feinberg, Kolesnikov, Han, Kwon, Chen, Chow, Zhu, Wei, Egyed, Cotruta, Giang, Kirk, Rao, Black, Babar, Lo, Moreira, Martins, Sanseviero, Gonzalez, Gleicher, Warkentin, Mirrokni, Senter, Collins, Barral, Ghahramani, Hadsell, Matias, Sculley, Petrov, Fiedel, Shazeer, Vinyals, Dean, Hassabis, Kavukcuoglu, Farabet, Buchatskaya, Alayrac, Anil, Dmitry, Lepikhin, Borgeaud, Bachem, Joulin, Andreev, Hardin, Dadashi, and Hussenot}]{gemmateam2025gemma3technicalreport}
Gemma Team, Aishwarya Kamath, Johan Ferret, Shreya Pathak, Nino Vieillard, Ramona Merhej, Sarah Perrin, Tatiana Matejovicova, Alexandre Ramé, Morgane Rivière, Louis Rouillard, Thomas Mesnard, Geoffrey Cideron, Jean bastien Grill, Sabela Ramos, Edouard Yvinec, Michelle Casbon, Etienne Pot, Ivo Penchev, and 197 others. 2025.
\newblock \href {https://arxiv.org/abs/2503.19786} {Gemma 3 technical report}.
\newblock \emph{Preprint}, arXiv:2503.19786.

\bibitem[{Wolf et~al.(2020)Wolf, Debut, Sanh, Chaumond, Delangue, Moi, Cistac, Rault, Louf, Funtowicz, Davison, Shleifer, von Platen, Ma, Jernite, Plu, Xu, Scao, Gugger, Drame, Lhoest, and Rush}]{wolf-etal-2020-transformers}
Thomas Wolf, Lysandre Debut, Victor Sanh, Julien Chaumond, Clement Delangue, Anthony Moi, Pierric Cistac, Tim Rault, Rémi Louf, Morgan Funtowicz, Joe Davison, Sam Shleifer, Patrick von Platen, Clara Ma, Yacine Jernite, Julien Plu, Canwen Xu, Teven~Le Scao, Sylvain Gugger, and 3 others. 2020.
\newblock Transformers: State-of-the-art natural language processing.
\newblock In \emph{Proc. of EMNLP: System Demonstrations}.

\bibitem[{Xia et~al.(2024)Xia, Yang, Dong, Wang, Li, Ge, Liu, Li, and Sui}]{xia2024unlockingefficiencylargelanguage}
Heming Xia, Zhe Yang, Qingxiu Dong, Peiyi Wang, Yongqi Li, Tao Ge, Tianyu Liu, Wenjie Li, and Zhifang Sui. 2024.
\newblock \href {https://arxiv.org/abs/2401.07851} {Unlocking efficiency in large language model inference: A comprehensive survey of speculative decoding}.
\newblock \emph{Preprint}, arXiv:2401.07851.

\bibitem[{Zhang et~al.(2025)Zhang, Xu, Liang, Chen, He, Wang, and Tu}]{zhang2024draft}
Ziyin Zhang, Jiahao Xu, Tian Liang, Xingyu Chen, Zhiwei He, Rui Wang, and Zhaopeng Tu. 2025.
\newblock \href {https://arxiv.org/abs/2411.18462} {Draft model knows when to stop: Self-verification speculative decoding for long-form generation}.
\newblock \emph{Preprint}, arXiv:2411.18462.

\bibitem[{Zheng et~al.(2023)Zheng, Chiang, Sheng, Zhuang, Wu, Zhuang, Lin, Li, Li, Xing, Zhang, Gonzalez, and Stoica}]{zheng2023judgingllmasajudgemtbenchchatbot}
Lianmin Zheng, Wei-Lin Chiang, Ying Sheng, Siyuan Zhuang, Zhanghao Wu, Yonghao Zhuang, Zi~Lin, Zhuohan Li, Dacheng Li, Eric~P. Xing, Hao Zhang, Joseph~E. Gonzalez, and Ion Stoica. 2023.
\newblock \href {https://arxiv.org/abs/2306.05685} {Judging llm-as-a-judge with mt-bench and chatbot arena}.
\newblock \emph{Preprint}, arXiv:2306.05685.

\end{thebibliography}

\clearpage
\appendix

\section{Appendix}

\subsection{Arm Algorithms}
\label{sec:arms}

Our bandits use five different algorithms as arms which are described briefly below. Let \( p(x_t \mid x_{t-1}, x_{t-2}, \ldots) \) denote the draft token distribution at timestep \( t \), and let \( \mathcal{H}(p) \) denote the entropy of this distribution.

\begin{itemize}
    \item \textbf{SVIP} \\
    \textit{Decision rule:}
    \[
    \sqrt{\mathcal{H}(p(x_t \mid x_{<t}))} > h
    \]
    where \( h \) is a fixed entropy threshold.

    \item \textbf{SVIP-Difference} \\
    \textit{Decision rule:}
    \begin{align*}
    \sqrt{\mathcal{H}(p(x_t \mid x_{<t}))} - \sqrt{\mathcal{H}(p(x_{t-1} \mid x_{<t-1}))} \\
    > h
    \end{align*}
    This captures a spike in uncertainty between two consecutive token steps.

    \item \textbf{Confidence} \\
    \textit{Decision rule:}
    \[
    p(x_t = \hat{x}_t \mid x_{<t}) < h
    \]
    where \( \hat{x}_t = \arg\max_x p(x \mid x_{<t}) \), and \( h \) is a confidence threshold.

    \item \textbf{Logit Margin} \\
    \textit{Decision rule:}
    \[
    p(x_t = \hat{x}_1 \mid x_{<t}) - p(x_t = \hat{x}_2 \mid x_{<t}) < h
    \]
    where \( \hat{x}_1 \) and \( \hat{x}_2 \) are the top-1 and top-2 predicted tokens respectively.

    \item \textbf{AdaEDL} \\
    \textit{Decision rule:}
    \[
    1 - \sqrt{\gamma \cdot \mathcal{H}(p(x_t \mid x_{<t}))} < \lambda_t
    \]
    \textit{Update rule:} After each full draft:
    \begin{align*}
        r_t &= \frac{n_{\text{acc}}}{n_{\text{drafted}}} \\
        \text{accept\_rate}_{t+1} &= \beta_1 \cdot \text{accept\_rate}_t \\ 
        &\phantom{= {}} + (1 - \beta_1) \cdot r_t \\
        \lambda_{t+1} &= \beta_2 \cdot \lambda_t \\
        &\phantom{= {}} + (1 - \beta_2) \cdot ( \lambda_t + \\
        &\phantom{= {}} +\epsilon \cdot \text{sign}(\alpha - r_t))
    \end{align*}
    
    where:
    \begin{itemize}
        \item \( n_{\text{acc}} \): Number of accepted tokens
        \item \( n_{\text{drafted}} \): Number of drafted tokens
        \item \( \alpha, \beta_1, \beta_2, \gamma, \epsilon \): Hyperparameters
    \end{itemize}

\end{itemize}

\subsection{Extra Ablations}
\label{sec:extra_ablations}

\subsubsection*{Adding More Possible Arms}
We also experiment with the possible arms the bandit can use. Specifically, for each dynamic speculation algorithm, we add several arms with different thresholds (e.g., entropy thresholds at 0.2, 0.4, and 0.6 for SVIP) rather than a single arm to see whether the opportunity to explore more arms results in better speedups. However, we find that the original setup with one threshold per dynamic speculation technique provides 12\% stronger speedups overall and adopt it for all future experiments.

\subsection{SpecBench Results}
\label{sec:specbench_results}
\begin{table}[t]
  \centering
  \caption{Evaluation of dynamic speculation methods on SpecBench across Llama-3, Gemma3, and OLMo-2 model families. \textbf{Bolded} results are the best and \textit{italicized} results are 2nd best.}
  \resizebox{\columnwidth}{!}{%
  \begin{tabular}{llcccc}
    \hline
    \textbf{Model} & \textbf{Method} & \textbf{Tuning Required?} & \textbf{\(m\)} & \textbf{\(\%\)} & \textbf{\(s\)} \\
    \hline
    \multirow{8}{*}{\shortstack[l]{Llama-3\\1B/70B}}
        & Static-6         & No & 3.49 & 0.58 & 1.00 \\
        & AdaEDL           & Yes & 2.50 & 0.88 & 0.96 \\
        & SVIP             & Yes & 3.37 & 0.85 & 1.07 \\
        & Max-Confidence   & Yes  & 4.70 & 0.80 & \textbf{1.18} \\
        & TapOut - Seq TS     & No  & 2.57 & 0.73 & 1.02 \\
        & TapOut - Seq UCB1   & No  & 4.70 & 0.81 & \textit{1.16} \\
        & TapOut - Token TS    & No  & 1.47 & 0.83 & 0.89 \\
        & TapOut - Token UCB1  & No  & 3.05 & 0.84 & 1.07 \\
    \hline

    \multirow{8}{*}{\shortstack[l]{Llama-3\\1B/8B}}
        & Static-6         & No & 3.29 & 0.55 & 1.00 \\
        & AdaEDL           & Yes & 1.53 & 0.86 & 1.08 \\
        & SVIP             & Yes & 2.18 & 0.85 & 0.68 \\
        & Max-Confidence   & Yes  & 2.77 & 0.81 & \textit{1.21} \\
        & TapOut - Seq TS     & No  & 1.00 & 0.80 & 0.90 \\
        & TapOut - Seq UCB1   & No  & 2.76 & 0.81 & \textbf{1.22} \\
        & TapOut - Token TS & No  & 2.23 & 0.82 & 1.13 \\
        & TapOut - Token UCB1   & No  & 1.19 & 0.79 & 0.92 \\
    \hline

    \multirow{8}{*}{\shortstack[l]{OLMo 2\\1B/32B}}
        & Static-6         & No & 1.93 & 0.32 & 1.00 \\
        & AdaEDL           & Yes & 0.67 & 0.67 & 1.10 \\
        & SVIP             & Yes & 1.15 & 0.68 & 1.23 \\
        & Max-Confidence   & Yes & 1.37 & 0.62 & \textbf{1.28} \\
        & TapOut - Seq TS     & No & 0.75 & 0.64 & 1.10 \\
        & TapOut - Seq UCB1   & No & 1.36 & 0.62 & 1.25 \\
        & TapOut - Token TS    & No & 0.75 & 0.65 & 1.11 \\
        & TapOut - Token UCB1  & No & 1.33 & 0.65 & \textit{1.26} \\
    \hline
    \multirow{8}{*}{\shortstack[l]{Gemma3 \\ 270M/27B}}
      & Static-6         & No  & 3.32 & 0.55 & 1.00 \\
      & AdaEDL           & Yes & 0.66 & 0.66 & 0.94 \\
      & SVIP             & Yes & 1.67 & 0.63 & \textit{1.04} \\
      & Max-Confidence   & Yes & 2.16 & 0.59 & 1.02 \\
      & TapOut - Seq TS     & No  & 0.73 & 0.63 & 0.93 \\
      & TapOut - Seq UCB1   & No  & 2.29 & 0.60 & \textbf{1.06} \\
      & TapOut - Token TS   & No  & 1.52 & 0.57 & 0.97 \\
      & TapOut - Token UCB1 & No  & 1.71 & 0.57 & 0.99 \\
    \hline
  \end{tabular}
  }
  \label{tab:results_specbench}
\end{table}
Table~\ref{tab:results_specbench} reports SpecBench~\citep{xia2024unlockingefficiencylargelanguage} outcomes across four model pairs: Llama-3.2 1B/3.1 70B and 3.2 1B/3.1 70B, OLMo-2 (1B/32B), and Gemma3 (270M/27B). The table lists whether tuning is required, the accepted length ($m$), the acceptance rate ($\%$), and the end-to-end speedup ($s$). Best results are bold and second-best are italic. Overall, sequence-level UCB1 is top-two for speedup across all model pairs while requiring no tuning.

\end{document}